\documentclass[11pt,a4paper]{article}
\usepackage[hyperref]{acl2019}
\usepackage{times}
\usepackage{latexsym}
\usepackage{graphicx}
\usepackage{wrapfig}

\usepackage{amsmath}
\usepackage{amssymb}
\usepackage{subcaption}

\usepackage{tikz}
\usetikzlibrary{shapes,arrows,positioning}

\usepackage[english]{babel}
\usepackage[T1]{fontenc}
\usepackage[utf8]{inputenc} 

\usepackage{url}

\usepackage{todonotes} 

\aclfinalcopy 

\title{Sparsity Emerges Naturally in Neural Language Models}

\author{Naomi Saphra  \and Adam Lopez \\
{\tt n.saphra@ed.ac.uk} \phantom{\and} {\tt alopez@ed.ac.uk} \\
  Institute for Language, Cognition, and Computation \\
  University of Edinburgh \\}

\begin{document}
\maketitle

\begin{abstract}
Concerns about interpretability, computational resources, and principled inductive priors have motivated efforts to engineer sparse  neural  models for NLP tasks. If sparsity is important for NLP, might well-trained neural models naturally become roughly sparse? Using the Taxi-Euclidean norm to measure sparsity, we find that frequent input words are associated with concentrated or sparse activations, while frequent target words are associated with dispersed activations but concentrated gradients. We find that  gradients associated with function words are more concentrated than the gradients of content words, even controlling for word frequency.
\end{abstract}

\section{Introduction}

Researchers in  NLP have long relied on engineering features to reflect the sparse  structures underlying language. Modern deep learning methods promised to  relegate this practice to history, but have not eliminated the interest in sparse modeling for NLP. Along with concerns about  computational resources \cite{chen2016compressing,narang2017exploring} and interpretability \cite{murphy2012learning,subramanian2018spine}, human intuitions continue to motivate sparse representations of language.  For example, some work   applies assumptions of sparsity to model latent hard categories such as syntactic dependencies \cite{pado2007dependency} or phonemes \cite{cotterell2018deep}. \citet{niculae2017regularized} found that a sparse attention mechanism outperformed dense methods on some NLP tasks; \citet{narang_exploring_2017} found sparsified versions of LMs that outperform dense originals. Attempts to engineer sparsity rest on an unstated assumption that it doesn’t arise naturally when neural models are learned. Is this true? 

Using a simple measure of sparsity, we analyze how it arises in different layers of a neural language model in relation to word frequency. We show that the sparsity of a word representation increases with exposure to that word during training.  We also find evidence of syntactic learning: gradient updates in backpropagation depend on whether a word's part of speech is open or closed class, even controlling for word frequency.

\section{Methods}

\paragraph{Language model.} Our LM is trained on a corpus of tokenized, lowercased English Wikipedia (70/10/20 train/dev/test split).  To reduce the  number of unique words (mostly names) in the corpus, we excluded any sentence with a word which appears fewer than 100 times. Those words which still appear fewer than 100 times after this filter are replaced with \verb=<UNK>=. The resulting training set is over 227 million tokens of around 19.5K types.
%

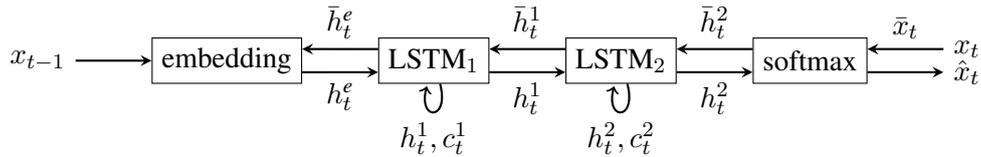
\begin{figure*}
\centering
\begin{tikzpicture}
	\node[rectangle, draw, minimum height=0.6cm, minimum width=1cm] (embedding) at (0, 0) {$\textrm{embedding}$};
	\node[rectangle, draw, right=of embedding, minimum height=0.6cm, minimum width=1cm] (RNN1) {$\textrm{LSTM}_1$};
	\node[rectangle, right=of RNN1, draw, minimum height=0.6cm, minimum width=1cm] (RNN2) {$\textrm{LSTM}_2$};
	\node[rectangle, right=of RNN2, draw, minimum height=0.6cm, minimum width=1cm] (softmax) {$\textrm{softmax}$};

\path (RNN1.east) -- (RNN1.north east) coordinate[pos=0.5] (RNN1_2_up);
\path (RNN2.west) -- (RNN2.north west) coordinate[pos=0.5] (RNN2_1_up);
\path (RNN1.south) -- (RNN1.south east) coordinate[pos=0.5] (RNN1_2_self);
\path (RNN1.south) -- (RNN1.south west) coordinate[pos=0.5] (RNN1_1_self);
\path (RNN2.south) -- (RNN2.south east) coordinate[pos=0.5] (RNN2_2_self);
\path (RNN2.south) -- (RNN2.south west) coordinate[pos=0.5] (RNN2_1_self);
\path (RNN1.east) -- (RNN1.south east) coordinate[pos=0.5] (RNN1_2_down);
\path (RNN2.west) -- (RNN2.south west) coordinate[pos=0.5] (RNN2_1_down);
\path (RNN1.west) -- (RNN1.north west) coordinate[pos=0.5] (RNN1_1_up);
\path (RNN2.east) -- (RNN2.north east) coordinate[pos=0.5] (RNN2_2_up);
\path (RNN1.west) -- (RNN1.south west) coordinate[pos=0.5] (RNN1_1_down);
\path (RNN2.east) -- (RNN2.south east) coordinate[pos=0.5] (RNN2_2_down);
\path (embedding.east) -- (embedding.north east) coordinate[pos=0.5] (embedding_up);
\path (embedding.east) -- (embedding.south east) coordinate[pos=0.5] (embedding_down);
\path (softmax.west) -- (softmax.north west) coordinate[pos=0.5] (softmax_1_up);
\path (softmax.west) -- (softmax.south west) coordinate[pos=0.5] (softmax_1_down);
\path (softmax.east) -- (softmax.north east) coordinate[pos=0.5] (softmax_2_up);
\path (softmax.east) -- (softmax.south east) coordinate[pos=0.5] (softmax_2_down);

	\node[left=of embedding] (X1) {$x_{t-1}$};
	\node[right=of softmax_2_up] (X2) {$x_t$};
	\node[right=of softmax_2_down] (X3) {$\hat{x}_t$};
	
\draw[-stealth, thick] (RNN1)  edge  [loop below] node[below] {$h_{t}^{1}, c_{t}^{1}$} ();
\draw[-stealth, thick] (RNN2)  edge  [loop below] node[below] {$h_{t}^{2}, c_{t}^{2}$} ();

    \draw[-stealth, thick] (X1) -- (embedding);
    \draw[-stealth, thick] (softmax_2_down) -- (X3) ;
    \draw[-stealth, thick] (X2) -- node[above] {$\bar{x}_{t}$} (softmax_2_up);
    \draw[-stealth, thick] (embedding_down) -- node[below]   {$h_{t}^{\emph{e}}$} (RNN1_1_down);
    \draw[-stealth, thick] (RNN1_1_up) -- node[above]   {$\bar{h}_{t}^{\emph{e}}$}  (embedding_up);
    \draw[-stealth, thick] (RNN1_2_down) -- node[below]  {$h_{t}^{1}$} (RNN2_1_down) ;
    \draw[-stealth, thick] (RNN2_1_up) -- node[above]  {$\bar{h}_{t}^{1}$} (RNN1_2_up) ;
    \draw[-stealth, thick] (RNN2_2_down) -- node[below] {$h_{t}^{2}$}  (softmax_1_down) ;
    \draw[-stealth, thick] (softmax_1_up) -- node[above] {$\bar{h}_{t}^{2}$}  (RNN2_2_up) ;
\end{tikzpicture}
\caption{LM architecture for target word distribution $\hat{x}_t$, showing gradient updates from observed word $x_t$.}
\label{fig:model}
\end{figure*}

We use a standard 2-layer LSTM LM trained with cross entropy loss for 50 epochs. 
The pipeline from input $x_{t-1}$ at time step $t-1$ to predicted output distribution $\hat{x}$ for time $t$ is described in Figure~\ref{fig:model}, illustrating intermediate activations $h_t^e$, $h_t^1$, and $h_t^2$. At training time, the network observes $x_t$ and backpropagates the gradient updates $\bar{h}_t^e$, $\bar{h}_t^1$,  $\bar{h}_t^2$, and $\bar{x}_t$.

The embeddings produced by the encoding layer are 200 units, and the recurrent layers have 200 hidden units each. The batch size is set to forty, the maximum sequence length to 35, and the dropout ratio to 0.2. The optimizer is standard SGD with clipped gradients at $\ell_2 = 0.25$, where the learning rate begins at 20 and is quartered whenever loss fails to improve. 

\paragraph{Measuring sparsity.}
We measure the sparsity of a vector $v$ using the reciprocal of the Taxicab-Euclidean norm ratio \citep{repetti2015euclid}. This measurement has a long history as a measurement of sparsity in natural settings \citep{zibulevsky2001blind,hoyer2004non,pham2017noise,yin2014ratio} and is formally defined as $\chi(v) = \|v \|_2 / \|v \|_1$.
The relationship between\begin{wrapfigure}{r}{0.15\textwidth}\vspace{-3mm}
\begin{center}
\includegraphics[scale=0.11]{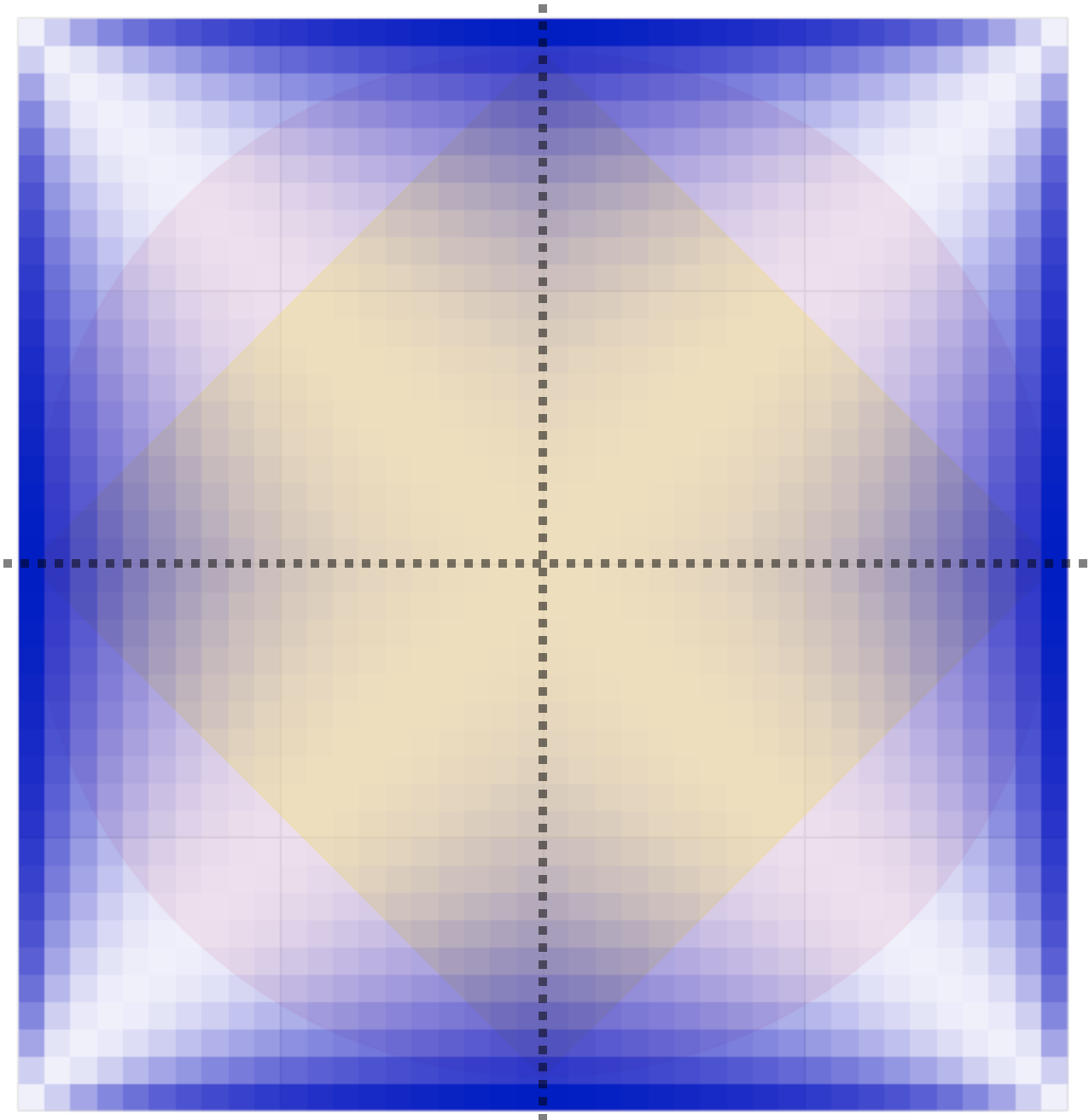}
\end{center}\vspace{-3mm}
\end{wrapfigure}
sparsity and this ratio is illustrated in two dimensions in the image on the right, in which darker blue regions are more concentrated. The pink circle shows the area where $\ell_2 \leq 1$ while the yellow diamond depicts $\ell_1 \leq 1$. For sparse vectors  $\langle 1, 0\rangle$ or $\langle 0, 1\rangle$, the norms are identical so $\chi$ is 1, its maximum. For a uniform vector like $\langle 1, 1\rangle$, $\chi$ is at its smallest. In general, $\chi(v)$ is higher when most elements of $v$ are close to 0; and lower when the elements are all similar in value.

\begin{figure}
\includegraphics[width=0.5\textwidth]{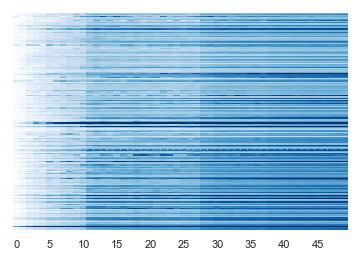}
\caption{Average sparsity $\chi(\bar{h_t^2})$ over all training epochs (x-axis), for target words $x_{t}$ occurring more than 100k times in training. Target words are sorted from most frequent (bottom) to least frequent (top).}
 \label{fig:rnn_out_concentration}
\end{figure}


\begin{figure*}
\begin{subfigure}{0.48\textwidth}
\includegraphics[width=\textwidth]{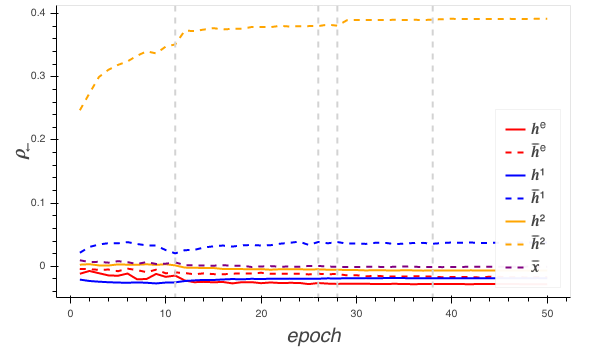}
\caption{$\rho_{\leftarrow}$ correlation with target word frequency} \label{fig:target_concentration}
\end{subfigure}
\begin{subfigure}{0.48\textwidth}
\includegraphics[width=\textwidth]{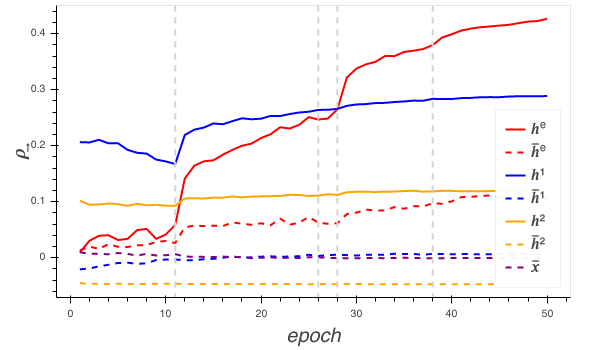}
\caption{$\rho_{\rightarrow}$ correlation with input word frequency} \label{fig:source_concentration}
\end{subfigure}
\caption{Correlation between mean sparsity of a word's representation and word frequency. Vertical dashed lines indicate when the optimizer has rescaled the step size.} 
\label{fig:concentration}
\end{figure*}

\section{Experiments}

Sparsity is closely related to the behavior of a model: If only a few units hold most of the mass of a representation, the \textit{activation} vector will be highly concentrated. If a neural network relies heavily on a small number of units in determining its predictions, the \textit{gradient} will be highly concentrated. A highly concentrated gradient is mainly modifying a few specific pathways. For example, it might modify a neuron associated with particular inputs like parentheses \cite{karpathy_visualizing_2015}, or properties like sentiment \cite{radford_learning_2017}.

\paragraph{Representations of Target Words.}
Our first experiments look at the relationship of sparsity to target word $x_t$. Gradient updates triggered by the target are often used to identify units that are relevant to a prediction~\citep{li_visualizing_2015}, and as shown in Figure~\ref{fig:rnn_out_concentration}, gradient sparsity increases with both the frequency of a word in the corpus and the overall training time. In other words, more exposure leads to sparser relevance. Because the sparsity of $\bar{h}^2$ increases with target word frequency,   we measure not sparsity itself but the Pearson correlation, over all words $w$, between word frequency and mean $\chi(h)$ over representations $h$ where $w$ is the target:
$$
\rho_{\leftarrow}(h) = \textrm{corr}_w(\mu_{t: x_t = w}(\chi(h_t)), \textrm{freq}(w))
$$ 
Here (Figure~\ref{fig:target_concentration}) we confirm that concentrated gradients are not a result  of concentrated activations, as activation sparsity $\chi(h^2)$ is not correlated with target word frequency.

The correlation is strong and increasing only for $\rho_{\leftarrow}(\bar{h}^2)$. The sparse structure being applied is therefore particular to the gradient passed from the softmax to the top LSTM layer, related to how a word interacts with its context.

\begin{figure}
\includegraphics[width=0.5\textwidth]{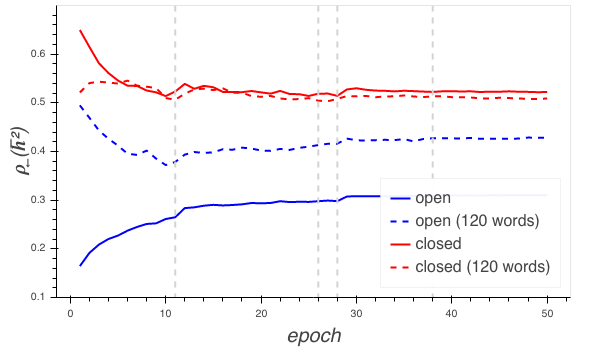}
\caption{$\rho_{\leftarrow}(\bar{h}^2)$, evaluated over vocabulary from open and closed classes of POS.}
 \label{fig:open_close}
\end{figure}

\paragraph{The Role of Part of Speech.}
Figure~\ref{fig:open_close} shows that $\rho_{\leftarrow}(\bar{h}^2)$ follows distinctly different trends for open POS classes\footnote{ADJ, ADV, INTJ, NOUN, PROPN, VERB} and closed classes\footnote{ADP, AUX, CCONJ, DET, PART, PRON, SCONJ}. 
To associate words to POS, we tagged our training corpus with spacy\footnote{https://spacy.io/}; we associate a word to a POS only if the majority (at least 100) of its occurrences are tagged with that POS.
We see that initially, frequent words from closed classes are highly concentrated, but soon stabilize, while frequent words from open classes continue to become more concentrated throughout training. Why?

Closed class words clearly signal POS. But open classes contain many ambiguous words, like ``report'', which can be a noun or verb. Open classes also contain many more words in general. We posit that early in training, closed classes reliably signal syntactic structure, and are essential for shaping network structure. But open classes are essential for  predicting specific words, so their importance in training continues to increase after part of speech tags are effectively learned. 

\begin{figure}
\includegraphics[width=0.5\textwidth]{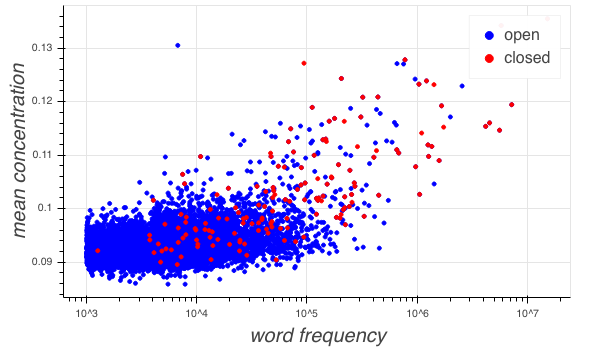}
\caption{Mean sparsity of $\chi(\bar{h}^{2})$ after 50 epochs, for words occurring more than 1k times in the train set.}
 \label{fig:scatterplot}
\end{figure}

The high sparsity of function word gradient may be surprising when compared with findings that content words have a greater influence on outputs \citep{kadar_representation_2016}. However, those findings were based on the impact on the vector representation of an entire sentence after omitting the word. \citet{khandelwal_sharp_2018} found that content words have a longer window during which they are relevant, which may explain the results of \citet{kadar_representation_2016}. Neither of these studies controlled for word frequency in their analyses contrasting content and function words,  but we believe this oversight is alleviated in our work by measuring correlations rather than raw magnitude. Because $\rho_{\leftarrow}(\bar{h}^2)$ is higher when evaluated over more frequent words, which also tend to be function words (see Figure~\ref{fig:scatterplot}), we further control for the effect of frequency by including a measurement of trends in a sample of 120 words each from open and closed classes (Figure~\ref{fig:open_close}). This sample was selected by sorting all open and closed class words by frequency, then choosing a range of each sorted list with a similar average frequency.

\paragraph{Representations of Input Words.} We next looked at the vector representations of each step in the word sequence as a representation of the input word $x_{t-1}$ that produced that step. We measure the correlation with input word frequency:
$$\rho_{\rightarrow}(h) = \textrm{corr}_w(\mu_{t: x_{t-1} = w}(\chi(h_t)), \textrm{freq}(w))$$

Here (Figure~\ref{fig:source_concentration}) we find that the view across training sheds some light on the learning process. While the lower recurrent layer quickly learns sparse representations of common input words, $\rho_{\rightarrow}(h^1)$ increases more slowly later in training and is eventually surpassed by $\rho_{\rightarrow}(h^\textrm{e})$, while gradient sparsity never becomes significantly correlated with word frequency.
\citet{li_understanding_2016} studied the activations of  feedforward networks in terms of  the importance of individual units  by erasing a particular dimension and measuring the difference in log likelihood of the target class. They found that importance  is concentrated into a small number of units at the lowest layers in a neural network, and is more dispersed at higher layers. Our findings suggest that  this effect may be a natural result of the sparsity of the activations at lower layers.

We relate the trajectory over training to the Information Bottleneck Hypothesis of \citet{shwartz-ziv_opening_2017}. This theory, connected to language model training by \citet{saphra2018understanding}, proposes that the earlier stages of training  are dedicated to  learning to effectively represent inputs, while later in training these representations are compressed and the optimizer  removes input  information extraneous to the task of predicting outputs. If extraneous information is encoded in specific  units, this compression would lead to the observed effect, in which  the first time the optimizer rescales the step size, it  begins an upward trend in $\rho_{\rightarrow}$ as  extraneous units are mitigated.

\section{Potential Explanations}

Why do common target words have such concentrated gradients with respect to the final LSTM layer?  
A tempting explanation is that the amount of information we have about common words offers high confidence and stabilizes most of the weights, leading to generally smaller gradients. If this were true, the denominator of sparsity, gradient $\ell_1$, should be  strongly anti-correlated with word frequency. In fact, it is only ever slightly anti-correlated (correlation $ > -.1$).  Furthermore, the sparsity of the softmax gradient $\chi(\bar{x})$ does not exhibit the strong correlation seen in $\chi(\bar{h}^2)$, so sparsity at the LSTM gradient is not a direct effect of sparse logits. 

However, the model could still be ``high confidence'' in terms of how it assigns blame for error during common events, even if it is barely more confident overall in its predictions. According to this hypothesis, a few specialized neurons might be responsible for the handling of such words.

Perhaps common words play a prototyping role that defines clusters of other words, and therefore have a larger impact on these clusters by acting as attractors within the representation space early on. Such a process would be similar to how humans acquire language by learning to use words like `dog' before similar but less prototypical words like `canine' \citep{rosch1999principles}. As a possible mechanism for prototyping with individual units, \citet{dalvi2019one} found that some neurons in a translation system specialized in particular word forms, such as verb inflection or comparative and superlative adjectives.  For example, a common comparative adjective like `better' might be used as a reliable signal to shape the handling of comparatives by triggering specialized units, while rarer words have representations that are more distributed according to a small collection of specific contexts. 

There may also be some other reason that common words interact more with specific substructures within the network. For example, it could be related to the use of context. Because rare words use more context than common words and content words use more context than function words \citep{khandelwal_sharp_2018}, the gradient associated with a common word would be focused on  interactions with the most recent words. This would lead common word gradients to be more concentrated.
    
It is  possible that  frequent words have sparse activations because  frequency  is learned as a feature and thus is counted by a few dimensions of   proportional magnitude, as posited by \citet{li_understanding_2016}. 

\section{Potential Applications}

Understanding where natural sparsity emerges in dense networks could be a useful guide in  deciding which layers we can apply sparsity constraints to without affecting  model performance, for the purpose of interpretability or efficiency. It might also explain why certain techniques are effective: for example, in some applications, summing representations together works quite well \cite{hill2016learning}. We hypothesize that this occurs when the summed representations are sparse so there is often little overlap. Understanding sparsity could help identify cases where such simple ensembling approaches are likely to be effective.

Future work may develop ways of manipulating the training regime, as in curriculum learning, to accelerate the concentration of common words or incorporating concentration into the training objective as a regularizer. 
We would also like to see how sparsity emerges in models designed for specific end tasks, and to see whether concentration is a useful measure for the information compression predicted by the Information Bottleneck.


\bibliography{emnlp2018}

\begin{thebibliography}{24}
\expandafter\ifx\csname natexlab\endcsname\relax\def\natexlab#1{#1}\fi

\bibitem[{Chen et~al.(2016)Chen, Mou, Xu, Li, and Jin}]{chen2016compressing}
Yunchuan Chen, Lili Mou, Yan Xu, Ge~Li, and Zhi Jin. 2016.
\newblock Compressing neural language models by sparse word representations.
\newblock \emph{arXiv preprint arXiv:1610.03950}.

\bibitem[{Cotterell and Eisner(2018)}]{cotterell2018deep}
Ryan Cotterell and Jason Eisner. 2018.
\newblock A deep generative model of vowel formant typology.
\newblock \emph{arXiv preprint arXiv:1807.02745}.

\bibitem[{Dalvi et~al.(2019)Dalvi, Durrani, Sajjad, Belinkov, Bau, and
  Glass}]{dalvi2019one}
Fahim Dalvi, Nadir Durrani, Hassan Sajjad, Yonatan Belinkov, Anthony Bau, and
  James Glass. 2019.
\newblock What is one grain of sand in the desert? analyzing individual neurons
  in deep nlp models.
\newblock In \emph{Proceedings of the AAAI Conference on Artificial
  Intelligence (AAAI)}.

\bibitem[{Hill et~al.(2016)Hill, Cho, Korhonen, and Bengio}]{hill2016learning}
Felix Hill, Kyunghyun Cho, Anna Korhonen, and Yoshua Bengio. 2016.
\newblock Learning to understand phrases by embedding the dictionary.
\newblock \emph{Transactions of the Association for Computational Linguistics},
  4:17--30.

\bibitem[{Hoyer(2004)}]{hoyer2004non}
Patrik~O Hoyer. 2004.
\newblock Non-negative matrix factorization with sparseness constraints.
\newblock \emph{Journal of machine learning research}, 5(Nov):1457--1469.

\bibitem[{Karpathy et~al.(2015)Karpathy, Johnson, and
  Fei-Fei}]{karpathy_visualizing_2015}
Andrej Karpathy, Justin Johnson, and Li~Fei-Fei. 2015.
\newblock Visualizing and {Understanding} {Recurrent} {Networks}.
\newblock \emph{arXiv:1506.02078 [cs]}.
\newblock ArXiv: 1506.02078.

\bibitem[{Khandelwal et~al.(2018)Khandelwal, He, Qi, and
  Jurafsky}]{khandelwal_sharp_2018}
Urvashi Khandelwal, He~He, Peng Qi, and Dan Jurafsky. 2018.
\newblock Sharp {Nearby}, {Fuzzy} {Far} {Away}: {How} {Neural} {Language}
  {Models} {Use} {Context}.
\newblock \emph{arXiv:1805.04623 [cs]}.
\newblock ArXiv: 1805.04623.

\bibitem[{{Kádár} et~al.(2016){Kádár}, {Chrupała}, and
  Alishahi}]{kadar_representation_2016}
{Ákos} {Kádár}, {Grzegorz} {Chrupała}, and Afra Alishahi. 2016.
\newblock Representation of linguistic form and function in recurrent neural
  networks.
\newblock \emph{arXiv:1602.08952 [cs]}.
\newblock ArXiv: 1602.08952.

\bibitem[{Li et~al.(2015)Li, Chen, Hovy, and Jurafsky}]{li_visualizing_2015}
Jiwei Li, Xinlei Chen, Eduard Hovy, and Dan Jurafsky. 2015.
\newblock Visualizing and understanding neural models in {NLP}.
\newblock \emph{arXiv preprint arXiv:1506.01066}.

\bibitem[{Li et~al.(2016)Li, Monroe, and Jurafsky}]{li_understanding_2016}
Jiwei Li, Will Monroe, and Dan Jurafsky. 2016.
\newblock Understanding neural networks through representation erasure.
\newblock \emph{arXiv preprint arXiv:1612.08220}.

\bibitem[{Murphy et~al.(2012)Murphy, Talukdar, and
  Mitchell}]{murphy2012learning}
Brian Murphy, Partha Talukdar, and Tom Mitchell. 2012.
\newblock Learning effective and interpretable semantic models using
  non-negative sparse embedding.
\newblock \emph{Proceedings of COLING 2012}, pages 1933--1950.

\bibitem[{Narang et~al.(2017{\natexlab{a}})Narang, Diamos, Sengupta, and
  Elsen}]{narang_exploring_2017}
Sharan Narang, Gregory Diamos, Shubho Sengupta, and Erich Elsen.
  2017{\natexlab{a}}.
\newblock Exploring {Sparsity} in {Recurrent} {Neural} {Networks}.
\newblock \emph{arXiv:1704.05119 [cs]}.
\newblock ArXiv: 1704.05119.

\bibitem[{Narang et~al.(2017{\natexlab{b}})Narang, Elsen, Diamos, and
  Sengupta}]{narang2017exploring}
Sharan Narang, Erich Elsen, Gregory Diamos, and Shubho Sengupta.
  2017{\natexlab{b}}.
\newblock Exploring sparsity in recurrent neural networks.
\newblock \emph{arXiv preprint arXiv:1704.05119}.

\bibitem[{Niculae and Blondel(2017)}]{niculae2017regularized}
Vlad Niculae and Mathieu Blondel. 2017.
\newblock A regularized framework for sparse and structured neural attention.
\newblock In \emph{Advances in Neural Information Processing Systems}, pages
  3338--3348.

\bibitem[{Pad{\'o} and Lapata(2007)}]{pado2007dependency}
Sebastian Pad{\'o} and Mirella Lapata. 2007.
\newblock Dependency-based construction of semantic space models.
\newblock \emph{Computational Linguistics}, 33(2):161--199.

\bibitem[{Pham et~al.(2017)Pham, Oudompheng, Mars, and Nicolas}]{pham2017noise}
Mai~Quyen Pham, Benoit Oudompheng, J{\'e}r{\^o}me~I Mars, and Barbara Nicolas.
  2017.
\newblock A noise-robust method with smoothed $\ell_1/\ell_2$ regularization
  for sparse moving-source mapping.
\newblock \emph{Signal Processing}, 135:96--106.

\bibitem[{Radford et~al.(2017)Radford, Jozefowicz, and
  Sutskever}]{radford_learning_2017}
Alec Radford, Rafal Jozefowicz, and Ilya Sutskever. 2017.
\newblock Learning to {Generate} {Reviews} and {Discovering} {Sentiment}.
\newblock \emph{arXiv:1704.01444 [cs]}.
\newblock ArXiv: 1704.01444.

\bibitem[{Repetti et~al.(2015)Repetti, Pham, Duval, Chouzenoux, and
  Pesquet}]{repetti2015euclid}
Audrey Repetti, Mai~Quyen Pham, Laurent Duval, Emilie Chouzenoux, and
  Jean-Christophe Pesquet. 2015.
\newblock Euclid in a taxicab: Sparse blind deconvolution with smoothed
  $\ell_1/\ell_2$ regularization.
\newblock \emph{IEEE Signal Processing Letters}, 22(5):539--543.

\bibitem[{Rosch(1999)}]{rosch1999principles}
Eleanor Rosch. 1999.
\newblock Principles of categorization.
\newblock \emph{Concepts: core readings}, 189.

\bibitem[{Saphra and Lopez(2018)}]{saphra2018understanding}
Naomi Saphra and Adam Lopez. 2018.
\newblock Understanding learning dynamics of language models with svcca.
\newblock \emph{arXiv preprint arXiv:1811.00225}.

\bibitem[{Shwartz-Ziv and Tishby(2017)}]{shwartz-ziv_opening_2017}
Ravid Shwartz-Ziv and Naftali Tishby. 2017.
\newblock Opening the {Black} {Box} of {Deep} {Neural} {Networks} via
  {Information}.
\newblock \emph{arXiv:1703.00810 [cs]}.
\newblock ArXiv: 1703.00810.

\bibitem[{Subramanian et~al.(2018)Subramanian, Pruthi, Jhamtani,
  Berg-Kirkpatrick, and Hovy}]{subramanian2018spine}
Anant Subramanian, Danish Pruthi, Harsh Jhamtani, Taylor Berg-Kirkpatrick, and
  Eduard Hovy. 2018.
\newblock Spine: Sparse interpretable neural embeddings.
\newblock In \emph{Thirty-Second AAAI Conference on Artificial Intelligence}.

\bibitem[{Yin et~al.(2014)Yin, Esser, and Xin}]{yin2014ratio}
Penghang Yin, Ernie Esser, and Jack Xin. 2014.
\newblock Ratio and difference of l1 and l2 norms and sparse representation
  with coherent dictionaries.
\newblock \emph{Commun. Inform. Systems}, 14(2):87--109.

\bibitem[{Zibulevsky and Pearlmutter(2001)}]{zibulevsky2001blind}
Michael Zibulevsky and Barak~A Pearlmutter. 2001.
\newblock Blind source separation by sparse decomposition in a signal
  dictionary.
\newblock \emph{Neural computation}, 13(4):863--882.

\end{thebibliography}
\bibliographystyle{acl_natbib_nourl}

\end{document}